\documentclass[10pt,letterpaper,oneside]{article}

\usepackage[T1]{fontenc}
\usepackage[utf8]{inputenc}
\usepackage{amsmath}     
\usepackage{newtxtext}   
\usepackage{newtxmath}   

\usepackage[letterpaper,margin=25mm,footskip=12mm]{geometry}
\usepackage{float}
\usepackage{needspace}
\usepackage{flushend}
\usepackage{graphicx}
\usepackage{booktabs}
\usepackage{caption}
\usepackage{xcolor}
\usepackage{titlesec}
\usepackage{ragged2e}
\usepackage[hidelinks]{hyperref}
\hypersetup{
  pdftitle={De-Risking Robotic Welding for In-Space Metal Joining Through Physics-Grounded Digital Twins},
  pdfauthor={Zhou Daniel Hao, Umer Masood Chaudry, Robert Shaw, Andrew Cheney, Nick Ludford, Mark Sims, Hongbiao Dong},
  pdfsubject={Physics-grounded modelling and simulated robotic control for in-space welding}
}

\titleformat{\section}{\normalsize\bfseries}{\thesection.}{0.5em}{}
\titleformat{\subsection}{\normalsize\itshape}{\thesubsection}{0.5em}{}
\titleformat{\subsubsection}{\normalsize\itshape}{\thesubsubsection}{0.5em}{}
\titlespacing*{\section}{0pt}{1.6ex plus .3ex}{0.6ex}
\titlespacing*{\subsection}{0pt}{1.2ex plus .2ex}{0.4ex}
\titlespacing*{\subsubsection}{0pt}{1.0ex plus .2ex}{0.3ex}

\graphicspath{{figures/}}
\begin{document}

\thispagestyle{plain}
\begin{center}
  {\Large\bfseries De-Risking Robotic Welding for In-Space Metal Joining\\
   Through Physics-Grounded Digital Twins\footnotemark}\\[8pt]
  {\bfseries
   Zhou Daniel Hao\textsuperscript{a,b}\textsuperscript{*},
   Umer Masood Chaudry\textsuperscript{a,b},
   Robert Shaw\textsuperscript{c},
   Andrew Cheney\textsuperscript{b},
   Nick Ludford\textsuperscript{c},
   Mark Sims\textsuperscript{b},
   Hongbiao Dong\textsuperscript{d}}
\end{center}
\footnotetext{An alternative version was accepted for oral presentation at the
77th International Astronautical Congress (IAC 2026). This work was supported
by the UK Space Agency under National Space Innovation Programme 2 (NSIP 2)
through the ISPARK project.}

\vspace{0pt}
{\centering
\textsuperscript{a}\,DANiLab, University of Leicester, Leicester, United Kingdom\\
\textsuperscript{b}\,School of Computing and Mathematical Sciences, University of
Leicester, Leicester, United Kingdom\\
\textsuperscript{c}\,TWI Ltd, United Kingdom\\
\textsuperscript{d}\,School of Metallurgy and Materials, University of
Birmingham, Birmingham, United Kingdom\\[3pt]
* Corresponding Author: d.hao@le.ac.uk\par}

\vspace{3pt}
\begin{center}\textbf{Abstract}\end{center}

Growing interest in in-space servicing, manufacturing, and assembly is renewing
attention on reliable metal joining processes that can support modular
construction, on-orbit repair, and future production workflows. Robotic welding
is a promising candidate, but controlled operation in reduced-gravity and
low-pressure environments remains insufficiently understood, particularly where
process physics, sensing, and control are strongly coupled and flight-like
testing opportunities are limited. To address this challenge, the UK Space
Agency-supported ISPARK (Intelligent SPace Arc-welding Robotic Kit) project is
developing a physics-grounded digital twin framework for structured process
de-risking. This paper evaluates a model-based prototype combining a
finite-volume thermal weld-pool model, Gaussian-process (GP) and neural-network
surrogates, and a simulated six-axis manipulator, with interfaces specified for
future electrical, thermal and visual measurements. The reduced model includes
solid--liquid phase change, effective thermocapillary transport and a
local-pressure parameter. Gravity is assessed separately through a
hydrostatic-to-capillary pressure ratio; it does not enter the thermal energy
equation. In repeated five-fold cross-validation, the GP gives
lower normalised mean absolute error than the neural network for all eight
thermal and weld-pool outputs. For ten deterministic boundary cases, 92.5\,\% of
80 solver responses lie within nominal 90\,\% GP predictive intervals. A
three-output query, including input scaling and predictive variance, takes
249\,$\mu$s. In a simulated UR5 L-shaped-seam study, a proportional
controller adjusts the synergic power-source setting as travel speed decreases
near the corner. Finite-volume re-evaluation shows that all 12 closed-loop
operating points and 9 of 12 open-loop points satisfy an illustrative
1.0--3.0\,mm root-side-width band. These simulation results suggest that the
framework can help identify candidate welding conditions and evaluate control
strategies before experimental testing. Further work should assess model
predictions through thermal-vacuum and targeted reduced-gravity tests and
evaluate the coupled controller through hardware-in-the-loop trials using a
robotic arm.

\vspace{2pt}

\vspace{2pt}
\begin{minipage}{\textwidth}
\textbf{Nomenclature}\\[-1pt]
{\footnotesize
\setlength{\tabcolsep}{0pt}
\renewcommand{\arraystretch}{0.86}
\begin{tabular}{@{}p{0.18\textwidth}@{\hspace{0.6em}}p{0.29\textwidth}@{\hspace{0.02\textwidth}}p{0.18\textwidth}@{\hspace{0.6em}}p{0.29\textwidth}@{}}
\textit{Symbol} & \textit{Meaning} & \textit{Symbol} & \textit{Meaning} \\
\midrule
$a_s,C_S$ & sulphur activity and concentration
  & $A,K,k_1,\Gamma_s,\Delta H^0$ & surface-adsorption constants \\
$f_\ell$ & liquid fraction
  & $\rho,c_p,L_f,L_v,t_s$ & material and sheet properties \\
$g/g_0$ & gravity level
  & $\gamma,\gamma_p$ & surface tension and reference value \\
$h$ & specific enthalpy
  & $\mathrm{d}\gamma/\mathrm{d}T$ & temperature coefficient of surface tension \\
$I,V,U$ & welding current, arc voltage and travel speed
  & $\eta,f_\mathrm{drop}$ & arc efficiency and droplet power fraction \\
$k_\mathrm{eff},\boldsymbol{\kappa}$ & effective and tensor conductivity
  & $\sigma_q,\sigma_j$ & heat-flux and current radii \\
$L_x,L_y,L_\mathrm{rem}$ & domain lengths
  & $\chi,\chi_\parallel,\chi_\perp$ & conductivity enhancements \\
$p_\infty,p_\mathrm{atm},p_\mathrm{sat}$ & local, atmospheric and saturation pressures
  & $M,R,\alpha_e,\dot m$ & evaporation variables \\
$\dot q_v,q_\mathrm{loss},q_\mathrm{fin}$ & heat sources and losses
  & $\varepsilon,\sigma$ & emissivity and Stefan--Boltzmann constant \\
$T,T_0,T_s,T_\ell,T_\mathrm{surf}$ & temperature states
  & $\nu,\sigma_f,\sigma_n,\delta_{\mathbf{x}\mathbf{x}'}$ & GP hyperparameters \\
$\mathbf{x},\ell_d,k(\mathbf{x},\mathbf{x}')$ & GP input, length scale and covariance
  & $K_p,e,\delta_i$ & controller gain, error and increment \\
$S,\Delta S$ & synergic power-source setting and rate limit
  & $a_\mathrm{lat},r_b,v_c$ & corner-motion variables \\
$W,\hat W,W^\star$ & width, prediction and target
  & $\Pi_\mathrm{hc}$ & hydrostatic-to-capillary pressure ratio \\
\bottomrule
\end{tabular}\par}
\end{minipage}

\vspace{0pt}
\textbf{Acronyms/Abbreviations}\\
{\footnotesize
CMT (Cold Metal Transfer);
GMAW (gas metal arc welding);
GP (Gaussian process);
GTAW (gas tungsten arc welding);
HIL (hardware-in-the-loop);
IoU (intersection over union);
ISAM (in-space servicing, assembly and manufacturing);
ISPARK (Intelligent SPace Arc-welding Robotic Kit);
MAE (mean absolute error);
ROS (Robot Operating System);
URDF (Unified Robot Description Format).\par}

\vspace{1pt}

\twocolumn

\section{Introduction}

In-space servicing, assembly and manufacturing (ISAM) could support structures
whose dimensions are not constrained by a single launch-vehicle fairing and
could allow spacecraft to be repaired or upgraded in orbit
\cite{malshe2023factories, sears2018impact}. Realising either depends on a
capability that remains largely unqualified for flight: permanent, load-bearing
metal joining. Mechanical fasteners and adhesives cover part of the design space,
but neither restores parent-metal continuity in a pressure boundary or a primary
structural member, and both introduce additional mass. Ground demonstrators have
examined high-degree-of-freedom
robotic assembly of structures too large for single-launch deployment
\cite{hao2019groundbased}.

\begin{figure}[!htb]
  \centering
  \includegraphics[width=\columnwidth]{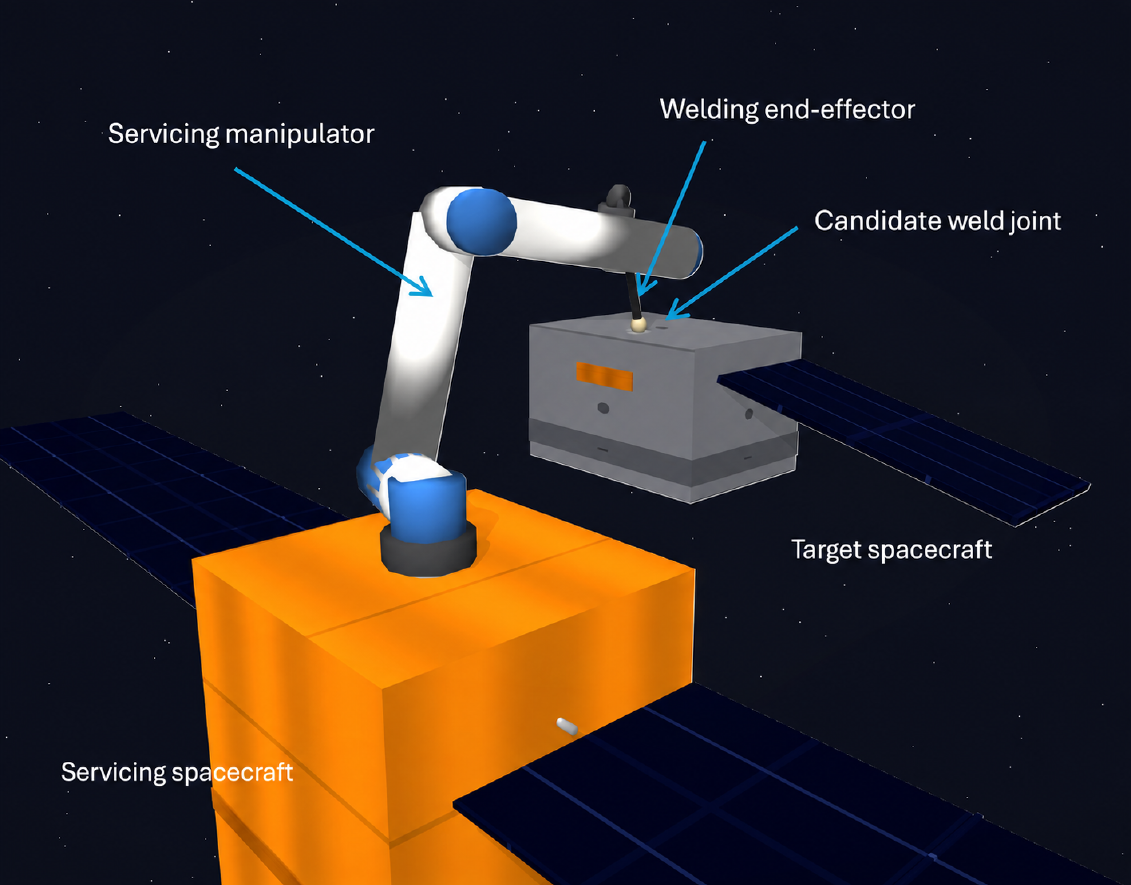}
  \caption{Concept for in-space robotic welding. A servicer-mounted
    six-axis manipulator positions the welding end-effector at a joint on the
    target spacecraft. Spacecraft geometry is illustrative. The manipulator
    simulation uses the UR5 kinematic description and Unified Robot Description
    Format (URDF) meshes published by
    Universal Robots~\cite{universalrobots2026description}.}
  \label{fig:concept}
\end{figure}

Figure~\ref{fig:concept} illustrates the representative on-orbit joining
scenario considered in this study.

Fusion welding is one candidate and has been tested in space. The Skylab M551
experiment produced electron-beam melts on 304 stainless steel, 2219-T87
aluminium and commercially pure tantalum in 1973, providing a rare in-space
fusion-welding dataset on structural alloys
\cite{busch1977m551}. Soviet
work established that an arc can be sustained in vacuum, while also documenting
its instabilities \cite{masubuchi1983feasibility}. Recent campaigns have
investigated laser beam welding in parabolic-flight and thermal-vacuum
demonstrations on AISI 316L
\cite{riffel2026pulsed, mcauley2026laser}.
Taken together, these studies do not report a qualified welding procedure or a
validated operating envelope spanning the gravity and pressure conditions
relevant to a mission.

Establishing that envelope experimentally is resource intensive. Reduced-gravity
access is limited to short parabolic-flight intervals, thermal-vacuum testing is
costly, and each run covers only a small part of the parameter space. The
coupled process physics therefore make exhaustive empirical mapping impractical.

A physics-based computational model can screen this operating space before
hardware tests, but a three-dimensional weld-pool calculation requires minutes
of computation per operating point. This cost prevents direct use at robotic
control rates and makes dense exploration of a five-dimensional design space
impractical. A faster surrogate model is therefore required.

This paper develops a reduced thermal weld-pool model parameterised by the CMT
operating condition and effective local-pressure parameter, compares Gaussian-process
and neural-network surrogates, and couples the GP surrogate to a
six-degree-of-freedom manipulator simulation. Gravity is sampled to check numerical
invariance and enters a separate hydrostatic-to-capillary pressure ratio, but is
absent from the implemented thermal energy equation. The
contributions are:

\begin{enumerate}\itemsep2pt
  \item A finite-volume thermal model applied here to 1\,mm 316L
        stainless-steel sheet, with explicit numerical convergence and
        field-quality criteria; extension to the 2\,mm coupon configuration is
        not evaluated in this paper.
  \item GP and neural-network surrogates evaluated by repeated cross-validation
        and deterministic boundary cases, with empirical interval coverage and
        measured inference latency.
  \item A model-based closed-loop study in which a controller adjusts the
        synergic power-source setting in response to a trajectory-imposed change
        in travel speed.
  \item A sensitivity analysis that separates conclusions supported within the
        reduced model from gravity-dependent free-surface and metal-transfer
        mechanisms that still require experiments and higher-fidelity models.
\end{enumerate}

\emph{Verification} concerns the numerical solution of the governing equations,
whereas \emph{validation} assesses agreement with physical observations.
This paper reports
numerical quality-control criteria, quantified
surrogate fidelity to the finite-volume model, and literature-informed parameter
priors. A formal grid-convergence study is still required before discretisation
uncertainty can be quantified. The paper does not claim
experimental validation: the targeted literature review identified no bead-
geometry dataset for 1--2\,mm 316L in the relevant travel-speed range, and
validation against the ISPARK coupons is unavailable in this study.

Section~\ref{sec:framework} sets out the framework architecture and its latency
budget. Section~\ref{sec:model} gives the weld-pool formulation, and
Section~\ref{sec:verification} its convergence and numerical-quality criteria.
Section~\ref{sec:priors} describes literature-informed parameter priors.
Section~\ref{sec:surrogate} reports surrogate fidelity and speed-up, and
Section~\ref{sec:robot} the model-based closed-loop study.
Section~\ref{sec:derisking} presents the model-based sensitivity results,
Section~\ref{sec:sensing} the sensing interfaces, and
Section~\ref{sec:roadmap} future experimental work and model development.

\section{The ISPARK digital-twin architecture}
\label{sec:framework}

The proposed architecture is organised into four layers with explicit interfaces. This
separation allows the finite-volume model to generate data offline while the
surrogate supplies predictions at control-loop rates.

\textit{Physics model.} A finite-volume thermal weld-pool model
(Section~\ref{sec:model}) solving enthalpy transport with solid--liquid phase
change and driven by a time-averaged arc heat input. The effective local-pressure
parameter enters the surface-loss model. Gravity is carried as a design
variable for an explicit invariance check and enters the separate
hydrostatic-to-capillary pressure ratio, but not the thermal energy equation. Cost is of order
$10^2$--$10^3$\,s per operating point.

\textit{Surrogate layer.} Gaussian-process and neural-network emulators trained
on solver sweeps (Section~\ref{sec:surrogate}), mapping a five-element process
and environment vector to weld-pool geometry, peak temperature and evaporation
rate. Each GP output has a nominal predictive interval. Cost is of order
$1.1\times10^{-4}$\,s for a single target and
$2.5\times10^{-4}$\,s for the three outputs queried by the control loop, measured on the
deployed predictor including input scaling and predictive variance.

\textit{Robotic coupling.} A six-degree-of-freedom manipulator with inverse
kinematics following the seam and querying the surrogate at each control step.
The controller adjusts the synergic power-source setting; travel speed is
prescribed by the motion plan (Section~\ref{sec:robot}). Related work combines
learned vision and manipulator trajectory generation in a spacecraft guidance,
navigation and control architecture for on-orbit manipulation
\cite{hao2021intelligent}.

\textit{Sensing interface.} The path by which measurements would update model
states and parameters
(Section~\ref{sec:sensing}). In this paper the interface is specified and the
assimilation structure defined, but the loop is not closed against hardware.

The thermal model generates surrogate training data offline. Within the online
control loop, the surrogate must provide sufficiently accurate predictions
within the available computation time. Measurement updates are part of the
proposed architecture but are not implemented in this study. The present
implementation is therefore a model-based control prototype within the proposed
architecture, not a hardware-connected digital twin.

The software interfaces could in principle support a different process model,
but transfer is not automatic. We infer from the process-specific diagnostics
and operating conditions used in recent laser-welding studies
\cite{riffel2026pulsed, mcauley2026laser} that a laser- or
electron-beam implementation would require its own heat-source model, inputs,
training data and validation, even if the surrogate and controller interfaces
were retained.

\begin{figure}[!htb]
  \centering
  \includegraphics[width=\columnwidth]{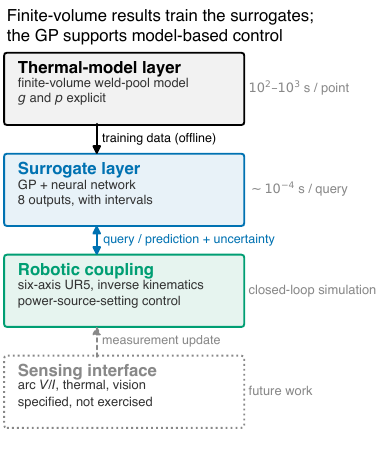}
  \caption{Proposed architecture for the ISPARK digital twin. The
    finite-volume weld-pool model generates training data offline for the
    Gaussian-process and neural-network surrogate layer; the surrogate then
    serves the robotic control loop online. Indicative costs annotated at right
    show the latency difference between the finite-volume model and the
    control-loop requirement. The
    sensing interface is drawn dotted because it is specified in this work but
    not tested with hardware.}
  \label{fig:framework}
\end{figure}

Figure~\ref{fig:framework} summarises the information flow and separates the
offline finite-volume calculation from the online robotic-control loop.

\section{Reduced thermal weld-pool model}
\label{sec:model}
\subsection{Formulation}

The model uses an \emph{equivalent heat source}, an established approach for
predicting thermal history and approximate weld-pool geometry in arc welding
\cite{goldak1984finite, cai2019heat}. The arc plasma, the electrode and the
metal-transfer cycle are not resolved; the process enters as a parameterised
heat input with a prescribed spatial distribution. The consequences of this
reduction are stated below and in Section~\ref{sec:verification}.

Energy is solved on a fixed Cartesian grid in a frame attached to the torch, with
the sheet translating at the travel speed $U$. A symmetry plane is imposed on the
weld centreline, so one half of the domain is modelled.

Specific enthalpy $h$ is advected to account explicitly for latent heat across
the melting front. For adjacent cells at the solidus and liquidus, the stated
material properties give an enthalpy difference of
$c_p \Delta T_m + L_f = 312$~kJ\,kg$^{-1}$. A sensible-heat-only approximation
gives $c_p \Delta T_m = 42$~kJ\,kg$^{-1}$, a factor of $7.4$ lower.
The scheme therefore advects $h$ in
conservative flux-divergence form
$\nabla\!\cdot\!(\mathbf{u}h)$, diffuses with face-averaged conductivity, and
inverts $h \rightarrow T$ exactly across the solid, mushy and liquid branches.
Internal face fluxes cancel in the discrete domain sum, leaving only boundary
fluxes. The non-conservative form $\mathbf{u}\!\cdot\!\nabla h$ omits the
$h\,\nabla\!\cdot\!\mathbf{u}$ term when the transport field has non-zero divergence.

The enthalpy--temperature relation carries the latent heat explicitly through the
liquid fraction $f_\ell$,
\begin{equation}
  \begin{aligned}
    h(T) &= c_p\,(T - T_0) + L_f\,f_\ell(T),\\
    f_\ell(T) &= \mathrm{clip}\!\left(
        \frac{T - T_s}{T_\ell - T_s},\,0,\,1\right),
  \end{aligned}
  \label{eq:enthalpy}
\end{equation}
and is inverted exactly, so no latent heat is lost at the phase boundary. The
transported form of the energy equation is then
\begin{equation}
  \frac{\partial h}{\partial t}
  + \nabla\!\cdot\!\left(\mathbf{u}\,h\right)
  = \frac{1}{\rho}\left[
      \nabla\!\cdot\!\left(\boldsymbol{\kappa}\,\nabla T\right)
      + \dot{q}_v \right],
  \label{eq:energy}
\end{equation}
with $\mathbf{u}$ the transport field in the torch frame, which reduces to the
sheet translation $U$ in the production configuration because momentum is not
solved. Consequently, gravity does not enter Eq.~\eqref{eq:energy}; its role in
the present implementation is limited to the diagnostic in
Eq.~\eqref{eq:pressure_ratio} and to an invariance check in the surrogate design.

The half-domain is $30\times6\times1$\,mm in the streamwise, transverse and
through-thickness directions. The unmeshed sheet extends from the 6\,mm model
boundary to the coupon half-width of 25\,mm and is represented by the finite-fin
condition below. The production grid contains $120\times48\times10$ cells,
giving spacings of $0.25\times0.125\times0.10$\,mm; the torch centre is 9\,mm
from the inlet. The explicit pseudo-transient march uses first-order upwind
enthalpy fluxes, a Courant number of 0.35, a maximum step of
$2.0\times10^{-4}$\,s and an additional diffusion-stability limit based on the
largest anisotropic effective diffusivity. A Rosenthal temperature field is
used only as the initial condition.

The energy equation carries a surface Gaussian arc flux of radius $\sigma_q$ and
a sub-surface volumetric source representing droplet enthalpy, splitting the net
power $\eta IV$ between them by a fraction $f_\mathrm{drop}$
\cite{cai2019heat}. Surface losses are convection, radiation
and evaporation. The reported calculations use $\eta=0.85$,
$\sigma_q=1.8$\,mm, $\sigma_j=2.0$\,mm and
$f_\mathrm{drop}=0.30$; the cylindrical droplet-energy source has a 1.0\,mm
radius and 0.6\,mm depth. These are literature-informed screening values, not
parameters fitted to the ISPARK coupons.

\subsection{Convective transport as an anisotropic effective conductivity}
\label{sec:convection}

Liquid-metal convection is represented by an enhanced anisotropic thermal
conductivity rather than by solving the momentum equations. This is a reduced
model of convective heat transport, not a prediction of the velocity field.
For the reference length and property scales, $\mathrm{Ra}\approx25$ and
$\mathrm{Ma}\approx1.2\times10^{3}$ at $1\,g$. These order-of-magnitude values
suggest that thermocapillary transport is more influential than buoyancy for the
reference case. The classical $\mathrm{Ra}_c\approx1700$ value is included only
as a shallow-layer reference and is not treated as a rigorous onset criterion
for the weld-pool geometry.

The net influence of thermocapillary flow on weld-pool geometry is directional.
Where
$\mathrm{d}\gamma/\mathrm{d}T < 0$ the surface flow is outward and the pool is
wide and shallow; where surfactant sulphur reverses the sign, the flow turns
inward and the pool deepens. An anisotropic conductivity represents that
directly,
\begin{equation}
  \begin{aligned}
    k_\mathrm{eff} &= k_s + \left(\chi\,k_\ell - k_s\right) f_\ell,\\
    \boldsymbol{\kappa} &= k_\mathrm{eff}
    \begin{cases}
      \mathrm{diag}(\chi_\parallel, \chi_\parallel, 1),
        & \tfrac{\mathrm{d}\gamma}{\mathrm{d}T} < 0,\\[2pt]
      \mathrm{diag}(1, 1, \chi_\perp),
        & \tfrac{\mathrm{d}\gamma}{\mathrm{d}T} > 0,
    \end{cases}
  \end{aligned}
  \label{eq:keff}
\end{equation}
the enhancement being applied laterally or through-thickness according to the
sign of $\mathrm{d}\gamma/\mathrm{d}T$ at the hottest surface point. The factors
$\chi=2.5$ and $\chi_\parallel=\chi_\perp=2.0$ are fixed in the reported
calculations. The first represents a baseline liquid-phase enhancement and the
second a directional multiplier, so the maximum directional conductivity is
$5k_\ell$. This two-stage construction approximates the net geometric effect of
thermocapillary transport but is not an independently calibrated decomposition.

This reduction cannot predict weld-pool velocity, recirculation structure or
mixing time. It also cannot resolve simultaneous inward and outward surface flow
when $\mathrm{d}\gamma/\mathrm{d}T$ changes sign within one pool; those quantities
require coupled momentum and free-surface calculations. Direct imaging of steel
arc-weld pools measured internal flow velocities of 0.1--0.5\,m\,s$^{-1}$ and
linked flow behaviour to the sign of $\mathrm{d}\gamma/\mathrm{d}T$
\cite{aucott2018revealing}.

\subsection{Sulphur-dependent surface tension}

The temperature coefficient of surface tension is not a material constant for
stainless steel: it depends on dissolved surfactant content, and its \emph{sign}
can invert. We adopt the Sahoo--Debroy--McNallan formulation
\cite{sahoo1988surface},
\begin{equation}
  \gamma(T) = \gamma_p - A(T-T_\ell) - R T \Gamma_s
              \ln\!\left(1 + K a_s\right),
  \label{eq:sahoo}
\end{equation}
with $K = k_1 \exp(-\Delta H^0/RT)$, $\Gamma_s$ the surface excess at
saturation and $a_s$ the dimensionless sulphur activity. The implementation
uses $\gamma_p=1.943$\,N\,m$^{-1}$, $A=4.3\times10^{-4}$\,N\,m$^{-1}$\,K$^{-1}$,
$\Gamma_s=1.3\times10^{-5}$\,mol\,m$^{-2}$, $k_1=0.00318$ and
$\Delta H^0=-1.88\times10^5$\,J\,mol$^{-1}$.
It
uses $a_s=f_s C_S$, where $C_S$ is entered as the numerical sulphur
concentration in wt\% and the activity coefficient is set to $f_s=1$ for this
screening study. Differentiating gives a coefficient that, for the adopted
coefficients, is positive near the liquidus above approximately
$C_S=0.005$\,wt\% and becomes negative at sufficiently high temperature. The
sign affects the
thermocapillary-flow direction: inward surface flow tends to produce a deeper,
narrower weld pool, whereas outward flow tends to produce a wider, shallower
pool \cite{aucott2018revealing}.

For the adopted parameter range, the transition temperature can fall inside the
weld-pool temperature range. The coefficient may therefore change sign between
the cooler periphery and hotter centre, producing counter-rotating flow regions
that the effective-conductivity model does not resolve
\cite{lu2008marangoni}.

We therefore treat sulphur concentration as an uncertain \emph{input} and sweep
$C_S=0.002$--$0.030$\,wt\%, rather than fixing a value. The upper bound is the
maximum specified for 316L sheet in ASTM A240/A240M; the lower bound is selected
for this screening study, not specified by the standard \cite{astm2023a240}.
Because the coupon heat analysis was unavailable, fixing
$\mathrm{d}\gamma/\mathrm{d}T$ to a nominal negative value would pre-determine
the predicted penetration response. Published measurements support the
relevance of the range: 316L plate reported at
0.021\,wt\% S \cite{liu2026enhancement} lies above the model's near-liquidus
sign-change value. This comparison does not establish the activity coefficient
or transition temperature for the ISPARK material.

\subsection{Reduced-pressure evaporation}

Evaporative mass flux is estimated with a Langmuir expression in which
$p_\infty$ is an imposed effective far-field partial pressure of the
evaporating species,
\begin{equation}
  \dot m = \alpha_e
  \left[\,p_\mathrm{sat}(T_\mathrm{surf}) - p_\infty\,\right]_{+}
  \sqrt{\frac{M}{2\pi R\,T_\mathrm{surf}}},
  \label{eq:langmuir}
\end{equation}
with $[\,\cdot\,]_{+}$ denoting the positive part and $\alpha_e = 0.82$ the
evaporation coefficient. Iron is used as the representative species, with
$M=0.0558$\,kg\,mol$^{-1}$, $L_v=6.34$\,MJ\,kg$^{-1}$ and
$p_\mathrm{sat}=p_\mathrm{atm}10^{6.121-18836/T}$. The resulting flux rises as
the imposed pressure falls and vanishes as it approaches the equilibrium vapour
pressure \cite{khan1984alloying}. The
free-surface balance removes convection, radiation and the associated enthalpy
flux,
\begin{equation}
  \begin{aligned}
    q_\mathrm{loss} &= h_c\,(T_\mathrm{surf} - T_0)
      + \varepsilon\sigma\!\left(T_\mathrm{surf}^4 - T_0^4\right)
      + \dot m\,L_v,\\
    h_c &= h_{c,0}\sqrt{p_\infty/p_\mathrm{atm}},
  \end{aligned}
  \label{eq:surfloss}
\end{equation}
where $h_{c,0}=15$\,W\,m$^{-2}$\,K$^{-1}$. The gas-side coefficient is assigned
the stated pressure scaling as a modelling assumption. In the vacuum limit,
radiation and evaporation remain as surface heat-loss terms. This treatment
provides the model's pressure dependence, while preferential loss of
manganese and chromium is a known compositional risk for vacuum welding of
stainless steels. The results are therefore reported as sensitivity to the
assumed coefficients, not as a calibrated evaporation prediction; primary
low-pressure evaporation coefficients for 316L were not available to this
study.

The imposed $p_\infty$ must not be interpreted automatically as the vacuum-
chamber pressure. A CMT process requires a local shielding-gas environment, and
the gas jet, arc plasma, metal-vapour plume and pressure-dependent arc
constriction are not resolved. The pressure results therefore describe the
specified local far-field parameterisation; mapping chamber pressure and gas
flow to that parameter requires a separate gas/arc model or measurement.

\subsection{Thin-sheet boundary conditions}

The reported calculations target full-penetration welding of unsupported 1\,mm
sheet. Both
sheet faces are therefore treated as free surfaces with radiation and
evaporation losses; no backing heat sink is included. Because full-penetration
pool equilibrium involves both free surfaces \cite{rybachuk2019weld}, root-side
and top-surface weld-pool widths are reported separately.

Transversely, the coupon extends beyond the modelled block to a fixture clamp.
Meshing that entire width would allocate cells to a thermally inactive region,
whereas an adiabatic truncation would prevent transverse heat loss. We therefore
apply the finite-fin condition
\begin{equation}
  q_\mathrm{fin} = k\,m\,\tanh\!\left(m L_\mathrm{rem}\right)(T - T_0),
  \qquad
  m = \sqrt{\frac{2\,h_\mathrm{eff}}{k\,t_s}},
  \label{eq:fin}
\end{equation}
which represents conduction through the unmeshed strip and heat loss from both
faces over the remaining width $L_\mathrm{rem}$. It approaches the
semi-infinite-fin limit when $mL_\mathrm{rem}\gg1$. A purely conductive flux
$k(T-T_0)/L_\mathrm{rem}$ omits surface losses from the strip and can make the
solution sensitive to the location of the computational boundary. Domain-independence
testing must confirm that the modelled boundary is sufficiently remote.

The hydrostatic-to-capillary pressure ratio is computed from the hydrostatic
head across the molten column and the capillary pressure that supports it,
\begin{equation}
  \Pi_\mathrm{hc} = \frac{\rho\,g\,t_s}{4\gamma/W}.
  \label{eq:pressure_ratio}
\end{equation}
This ratio compares hydrostatic and capillary pressure only. It excludes arc
pressure, free-surface deformation and dynamic metal transfer, and therefore is
not a stand-alone predictor of burn-through or melt-through.

\subsection{Limitations}

The free surface is flat (rigid-lid): weld reinforcement and root profile are not
resolved. Quantifying the effect of this assumption would require a free-surface
(volume-of-fluid) reference calculation, which is outside the scope of this
paper and is identified as the highest-priority model extension in
Section~\ref{sec:roadmap}. The model is driven by
time-averaged current and voltage and does not resolve the controlled dip-transfer
cycle of CMT or a pulsed-current waveform; the process enters through
$\eta$, $\sigma_q$ and $f_\mathrm{drop}$, which are fixed in the reported
calculations. Droplet detachment, arc-plasma coupling and
solidification microstructure are outside scope. Higher-fidelity work requires
linked thermal, fluid-flow, interfacial and solidification models supported by
experimental inputs and validation evidence \cite{tong2013multiscale}.

\section{Numerical convergence and quality control}
\label{sec:verification}

This section reports convergence and numerical-quality checks. In the standard
verification, validation and uncertainty-quantification terminology,
verification assesses implementation of the mathematical model, whereas
validation assesses its representation of the physical system. The present
work does not include experimental validation, and a formal grid-convergence
study is still outstanding.

\subsection{Convergence criterion}

Each case is marched pseudo-transiently to a quasi-steady state in the torch
frame. Convergence is accepted only when the following four conditions are
satisfied simultaneously:

\begin{enumerate}\itemsep2pt
  \item \textit{Drift.} Liquid volume and peak temperature must change by less than
        $2\times10^{-3}$ across six checks, each separated by 200 time steps,
        rather than between consecutive checks.
  \item \textit{Storage.} The rate of change of stored enthalpy, including the
        latent contribution, must fall below 4\,\% of the net input power.
  \item \textit{Balance.} The energy budget --- surface losses from both faces,
        advected enthalpy at the outlet, conduction into the transverse fin
        boundary and upstream into the inlet, plus storage --- must close on the
        net input power to within 8\,\%.
  \item \textit{Field quality.} The temperature maximum must lie on the symmetry
        plane to within one cell, the transverse surface profile must decay
        monotonically outboard of it, and the largest temperature difference
        between adjacent cells that are \emph{both} molten must remain below
        0.50 times the peak superheat.
\end{enumerate}

The final design comprised 60 Latin-hypercube cases and ten deterministic
boundary cases. All cases met the drift, storage and energy-balance thresholds;
two Latin-hypercube cases failed the local smoothness criterion and were
excluded. The surrogate was therefore trained and cross-validated on 58 cases,
with all ten deterministic boundary cases reserved for the separate assessment.

The drift criterion alone was insufficient in an early sweep: the reported weld
pool was nearly stationary while the surrounding sheet continued to absorb
heat, because the stop time had been sized on the advective residence time
$2.5L_x/U$ while the slowest mode in the domain is transverse diffusion,
$L_y^2/\alpha$, which is several times longer. Those cases reported drift
residuals of order $10^{-5}$ while
about a fifth of the input power was unaccounted for. Adding the storage term
converts that shortfall from an unexplained discrepancy into a measured
transient, distinguishing incomplete convergence from an omitted sink term.

The stop time is consequently driven by the diffusion time rather than the
residence time. Because that time scales as $L_y^2$, the modelled half-width is
an important determinant of computational cost, and the fin-conduction boundary
condition of Section~\ref{sec:model} permits this reduction. A formal
domain-independence result remains to be reported.

The fourth condition is independent of the integral checks. A conservative
discretisation can close the global energy balance while retaining a
non-physical local field. In an intermediate solver version, the energy budget
closed to $0.16\%$ of the net input power and every attempted case in a
70-case sweep
satisfied the drift and storage tests, while the temperature field differed by
approximately $1400$~K between adjacent cells and its maximum lay
$0.44$~mm off the symmetry plane, where a source centred on that plane cannot
place it. The cause was a transport field with non-zero divergence: written in
conservative flux form, the advection operator redistributed enthalpy between
neighbouring cells while conserving the global total exactly, which is precisely
why the global energy-balance check did not detect the defect. Field-level criteria are
therefore applied in addition to the global balance.

Cases in which any numerical safeguard remains active at termination, including
the imposed 3086\,K boiling-temperature cap or transport-field clipping, are
recorded and excluded from surrogate training because they do not represent the
unmodified governing model.

\subsection{Outstanding verification studies}

The 68 retained results satisfy the convergence and field-quality criteria above, but
they do not constitute a complete numerical-verification record. The remaining
studies are grid refinement with an observed order and Grid Convergence Index,
domain independence with respect to modelled half-width, initial-condition
independence, and a tabulated steady-state energy balance. Until these studies
are complete, discretisation uncertainty is not propagated through the
surrogate and the reported predictive intervals quantify surrogate uncertainty
only.

\section{Literature-informed parameter priors}
\label{sec:priors}

The targeted review identified no directly matching experimental dataset for
the present sheet thickness, material and operating range. The uncertain parameters are therefore
assigned literature-informed priors rather than fitted to ISPARK measurements.
Any future calibration could follow the Kennedy--O'Hagan structure
\cite{kennedy2001bayesian}, in which an observation is
represented by a simulator response, a model-discrepancy term and measurement
error. This study specifies priors only; no posterior calibration or
model-discrepancy estimate is reported.

A targeted literature corpus provided 41 quantitative records used to inform
parameter ranges. The closest accessible comparators
include autogenous gas tungsten arc welding (GTAW) bead-on-plate trials on 316L at
100\,A and 5\,mm\,s$^{-1}$ \cite{abdulkadir2021investigation} and a Taguchi L9
orthogonal array with ER316L metal-cored wire
\cite{wang2024preliminary}.

Four parameter groups are assigned literature-informed prior ranges:

\textit{Arc efficiency} $\eta$, with a prior range of 0.65--0.87 based on
calorimetry \cite{pepe2011measuring}. An
experimentally calibrated value for the ISPARK power-source programme is not
available in this study.

\textit{Sulphur-dependent surface tension}, through the concentration-to-activity
approximation in Eq.~\ref{eq:sahoo}, swept over
$C_S=0.002$--$0.030$\,wt\% rather than fixed.

\textit{Heat-flux and current-density distribution radii} $\sigma_q$ and
$\sigma_j$, with prior scales of 1--3\,mm
\cite{cai2019heat}, considered jointly with the droplet power
fraction $f_\mathrm{drop}$.

\textit{Effective-conductivity factors} $\chi$, $\chi_\parallel$ and
$\chi_\perp$, which represent unresolved liquid-metal convective heat transport
in the reduced thermal model \cite{hu2007heatb}.

Condensed-phase properties are fixed from a standard compilation
\cite{mills2002recommended} rather
than calibrated. The implemented values are
$\rho=7400$\,kg\,m$^{-3}$, $c_p=700$\,J\,kg$^{-1}$\,K$^{-1}$,
$k_s=22$ and $k_\ell=30$\,W\,m$^{-1}$\,K$^{-1}$,
$T_s=1663$\,K, $T_\ell=1723$\,K, $L_f=270$\,kJ\,kg$^{-1}$ and
$\varepsilon=0.4$. These fixed screening values are another source of model
discrepancy that is not included in the GP intervals.

Three evidence gaps bound the interpretation. First, the extracted records do
not provide a direct validation dataset for 1--2\,mm 316L over the present
travel-speed range. Second, no directly applicable burn-through threshold was
identified for unsupported thin stainless-steel sheet. Third, primary
low-pressure evaporation coefficients for 316L were unavailable, so evaporation
is treated as a sensitivity rather than a validated prediction. Existing
measurements establish the influence of sulphur and silicon on the surface
tension of 316L; additional measurements spanning controlled sulphur levels and
welding-relevant thermal and atmospheric conditions would be particularly
valuable \cite{grant1992surface}.

\section{Surrogate modelling and benchmark}
\label{sec:surrogate}

\subsection{Design and construction}

The surrogate maps a five-variable design --- synergic power-source setting,
travel speed, gravity level, the effective local-pressure parameter and sulphur
concentration --- to eight outputs:
top-surface weld-pool width, root-side width, penetration depth, pool length,
peak temperature, elongation, evaporation rate and width-to-depth aspect ratio.
Gravity and pressure enter as $\log_{10}(g/g_0)$ and
$\log_{10}(p_\infty/1\,\mathrm{Pa})$. Gravity spans $10^{-6}$--$1g$ and is
included to check numerical invariance, although it is inactive in the thermal
equations; pressure spans $10^{-3}$--101325\,Pa and is active through the
surface-loss assumptions.

The Latin-hypercube bounds are power-source setting 1--9, travel speed
6--20\,mm\,s$^{-1}$, $g/g_0=10^{-6}$--1,
$p_\infty=10^{-3}$--101325\,Pa and
$C_S=0.002$--0.030\,wt\%. Cases are retained only when the nominal net heat
input lies between 0.025 and 0.090\,kJ\,mm$^{-1}$. After this filter, the
accepted design spans settings 1.37--8.31 and travel speeds
8.03--19.97\,mm\,s$^{-1}$.

Welding current and arc voltage are not sampled independently. For the synergic
power-source programme they vary with wire-feed speed along a measured operating
line. A single setting interpolated along that line therefore replaces three
correlated inputs and avoids combinations unavailable from the power source.

The Gaussian process uses a Mat\'ern kernel with $\nu=5/2$ and automatic
relevance determination, one length scale $\ell_d$ per input,
\begin{equation}
  \begin{aligned}
    k(\mathbf{x},\mathbf{x}') &= \sigma_f^2\,k_{5/2}(r)
      + \sigma_n^2\,\delta_{\mathbf{x}\mathbf{x}'},\\
    r^2 &= \sum_{d=1}^{5}
      \frac{\left(x_d - x'_d\right)^2}{\ell_d^{2}},
  \end{aligned}
  \label{eq:kernel}
\end{equation}
chosen over the smoother squared-exponential default because the response near a
penetration threshold can contain a sharp change in slope. After input
standardisation, the fitted $\ell_d$ values are used as model-based sensitivity
indicators in Section~\ref{sec:derisking}; a length scale at its upper bound is
interpreted only as weak resolved dependence on that input over the sampled
domain, not as evidence of general physical irrelevance.

Training data come from a Latin-hypercube design over the five inputs
\cite{sacks1989design}, augmented with deterministic boundary cases held out of the
random design. Cases that failed, that did not satisfy the convergence criterion
of Section~\ref{sec:verification}, or in which a numerical safeguard remained
active at termination are excluded from training so that the surrogate is not
fitted to numerically limited fields.

Two surrogate families are compared. The GP is fitted independently to each
output using three optimiser restarts and random seed 0; the constant, length-
scale and white-noise bounds are $10^{-3}$--$10^3$,
$10^{-2}$--$10^4$ and $10^{-12}$--$10^1$, respectively. The comparison model is a
feed-forward multilayer perceptron with two
hidden layers of 64 neurons, hyperbolic-tangent activation,
standardised inputs and outputs, early stopping and random seed 0
\cite{hemmasian2023surrogate}. Separate output models provide
output-specific GP predictive intervals.

\subsection{Prediction accuracy and computational cost}

Interpolation accuracy is assessed by four repeats of five-fold cross-validation
and reported per output as mean absolute error (MAE), both in physical units and
as a fraction of the sampled output range, together with $R^2$. Ten
deterministic boundary cases are held out as a separate test. For the GP,
\emph{interval coverage} is the fraction of finite-volume responses within the
nominal 90\% predictive interval. Coverage is reported alongside point error
to assess predictive uncertainty; the control law uses the predicted mean.

Across the 80 boundary-case responses, 74 lie within the nominal 90\,\%
interval. Coverage is 100\,\% for six outputs, 80\,\% for evaporation rate and
60\,\% for root-side width. The overall coverage of 92.5\,\% therefore does not
establish calibrated uncertainty for every output.

\begin{figure*}[!t]
  \centering
  \includegraphics[width=\textwidth]{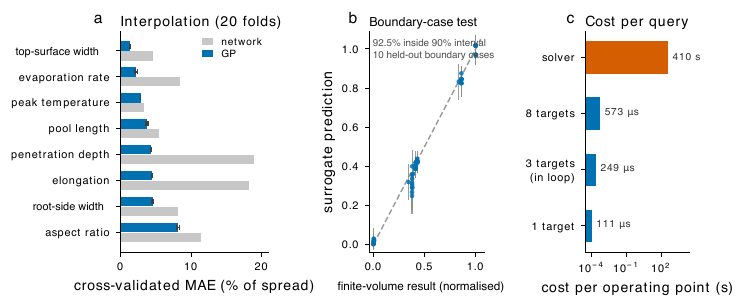}
  \caption{Surrogate accuracy and cost. (a) Mean absolute error normalised by
    output range in four repeats of five-fold cross-validation; bars show
    fold-to-fold standard deviation. (b) Ten boundary cases held out of the
    Latin hypercube; $92.5\,\%$ of 80 responses lie within nominal $90\,\%$ GP
    intervals and median error is $1.64\,\%$ of the output range. (c) Measured cost per
    operating point: the median of all 70 attempted solver cases and deployed
    surrogate queries including scaling and predictive variance. The
    three-output control query gives a speed-up of $1.6\times10^{6}$.}
  \label{fig:surrogate_benchmark}
\end{figure*}

Speed is reported as measured finite-volume wall time per operating point and
measured surrogate latency per query, including input scaling and GP predictive
variance. Single-output, three-output and eight-output queries are timed
separately. Figure~\ref{fig:surrogate_benchmark} summarises the accuracy,
boundary-case coverage and measured computational cost; the timing values are
specific to the reported implementation and execution platform.

\subsection{Weld-pool shape reconstruction from scalar outputs}
\label{sec:shape}

The surrogate predicts scalar weld-pool dimensions rather than the temperature
field. At the assumed reference condition, a transverse trapezoid formed from
top-surface width, root-side width and depth gives
$\mathrm{IoU}=0.93$ against the finite-volume liquidus boundary. This result is
specific to the reported 1\,mm case, whose pool walls are nearly vertical.

In plan view, a length--width ellipse anchored at the torch gives only
$\mathrm{IoU}=0.68$, whereas optimising its axial position raises IoU to $0.95$.
The liquidus contour extends about $0.7$~mm ahead of the torch, so the dominant
error is position rather than elliptical shape. Depth and root-side width retain
the information used by the controller, but a general shape model would require
additional outputs; adding the leading-edge offset would require regenerating
the training sweep.

\begin{figure}[!htb]
  \centering
  \includegraphics[width=\columnwidth]{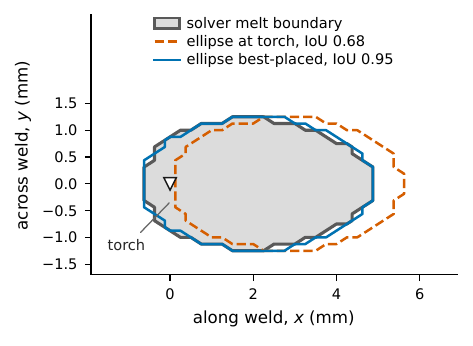}
  \caption{Plan-view liquidus boundary at the assumed reference condition
    ($10^{-6}\,g$, $10^{-3}$~Pa). The finite-volume contour is grey. An ellipse
    from surrogate length and width gives $\mathrm{IoU}=0.68$ at the torch
    (dashed) and $0.95$ after axial placement (solid), showing that position
    dominates the error. Blockiness reflects the $0.25\times0.125$~mm cells.}
  \label{fig:shape_recon}
\end{figure}

Figure~\ref{fig:shape_recon} compares the finite-volume liquidus boundary with
the two scalar reconstructions at the assumed reference condition.

\section{Closed-loop robotic coupling}
\label{sec:robot}

\subsection{Computational requirements for control}

A welding controller must update on the timescale of torch motion. The weld-pool
model of Section~\ref{sec:model} requires of order $10^2$--$10^3$\,s
per operating point, so closing a loop on it directly would run at a real-time
factor of order $10^{-5}$. At the measured median solver time, evaluating all
217 control steps for the 50\,mm seam would take approximately 25\,h. Using the
467.6\,s mean measured for the 24 finite-volume operating-point evaluations
below gives the more conservative estimate of 28\,h. The finite-volume model is
therefore retained as an offline reference rather than used directly for control.

\subsection{Implementation}

A six-axis UR5 manipulator is simulated with the standard ROS
\texttt{ur\_description} link parameters and full-pose inverse kinematics
\cite{universalrobots2026description}. The tool is a $220$\,mm torch held at a $15^\circ$ drag
angle referenced to the direction of travel, at a nominal
contact-tip-to-work distance of $12$\,mm. The path is an L-seam of two $25$\,mm
legs --- $50$\,mm total, matching the coupon weld length --- joined by a
$1.6$\,mm circular blend, so that the geometry and the speed profile describe the
same motion. The fixed-base inverse-kinematics result is independent of the
assigned gravity vector; the assumed microgravity value is retained only for
configuration consistency.

The corner introduces a speed change determined by the trajectory geometry.
A lateral-acceleration limit $a_\mathrm{lat}$ caps the speed through a blend of
radius $r_b$ at
\begin{equation}
  v_c = \sqrt{a_\mathrm{lat}\,r_b},
  \label{eq:vcorner}
\end{equation}
which for $a_\mathrm{lat}=0.05$\,m\,s$^{-2}$ and $r_b=1.6$\,mm gives
$v_c = 8.94$\,mm\,s$^{-1}$ against a nominal $12.0$\,mm\,s$^{-1}$, ramped over
$6$\,mm either side. Because heat input per unit length varies as $\eta IV/U$,
that slowdown raises it by a third at fixed machine settings, causing the
surrogate-predicted root-side width to exceed the illustrative upper limit. The
blend radius was selected so that $v_c$ remains inside the
$8.03$--$19.97$\,mm\,s$^{-1}$ travel-speed range the surrogate was trained over,
and the implementation checks this bound at run time to prevent surrogate
extrapolation.

Travel speed is prescribed by the motion plan and is identical in the open- and
closed-loop runs. It acts as the disturbance input, while the manipulated
variable is the synergic power-source setting. At the corner, the prescribed
speed is already limited by the manipulator's lateral-acceleration constraint;
the controller therefore changes the welding condition rather than the motion.

The controller adjusts the synergic power-source setting $S$ in proportion to
the error between the target width and the surrogate's predicted root-side width
$\hat W$, subject to a rate limit and configured setting bounds,
\begin{equation}
  \begin{aligned}
    \delta_i &= \mathrm{clip}\big(
        K_p\,[\,W^\star - \hat W_i\,],\, -\Delta S,\, +\Delta S\big),\\
    S_{i+1} &= \mathrm{clip}\big(
        S_i + \delta_i,\; S_\mathrm{min},\, S_\mathrm{max}\big),
  \end{aligned}
  \label{eq:control}
\end{equation}
with $W^\star = 2.0$\,mm, $K_p = 0.28$ setting units per millimetre,
$\Delta S = 0.25$ per step and $S \in [2,9]$. The model-based loop uses a
quasi-steady algebraic process map and therefore excludes physical thermal and
metal-transfer dynamics. Away from the rate and range limits, its local error
recursion is
$e \leftarrow \left(1 - K_p\,\partial W/\partial S\right) e$, stable for
\begin{equation}
  \left|\,1 - K_p\,\frac{\partial W}{\partial S}\,\right| < 1 .
  \label{eq:stability}
\end{equation}
The gain is selected from the local surrogate sensitivity:
$\partial W/\partial S = 2.95$\,mm per setting unit at nominal speed and
$1.89$\,mm per setting unit at the corner. Thus $K_p=0.28$ gives local error
multipliers of $0.17$ and $0.47$, respectively. These values satisfy
Eq.~\eqref{eq:stability} for the unsaturated algebraic model; they do not
establish stability for the future transient physical system.

Each control step issues one surrogate query, and the setting in force is the one
computed from the previous step's prediction. A proportional law is used so that
the contribution of prediction latency and error can be assessed without a more
complex controller. Each commanded change is logged with the prediction that
triggered it.

Reported metrics are the fraction of seam length inside the illustrative
root-side-width band, the peak excursion beyond it, the real-time factor, and the mean
inverse-kinematics and surrogate-query latencies per step.

The open- and closed-loop simulations execute at 52.7 and 53.6 times real time,
respectively. Their mean inverse-kinematics latencies are 0.112 and 0.111\,ms per
step, and their corresponding three-output surrogate-query latencies are 0.248
and 0.249\,ms per step. These wall-clock results are hardware-specific and do
not include communication with a physical robot controller.

\subsection{Scope}

The simulated control loop couples surrogate predictions and power-source
commands but excludes robot--structure dynamics. Manipulator reaction loads
could deform the supporting structure, alter contact-tip-to-work distance and
affect arc stability. Modelling this interaction requires a further extension.

A second limitation is that the controller acts on a surrogate prediction and
the same surrogate is used to evaluate performance along the full seam. That comparison alone
assesses the control law and latency budget, not performance against an
independent physical system. The finite-volume cross-check below reduces this
circularity but is still model-to-model evidence.

\subsection{Finite-volume-model assessment of commanded operating points}
\label{sec:sil}

The operating points commanded by each loop were re-evaluated with the
finite-volume model and scored against those predictions rather than the
surrogate's own outputs. This is a higher-fidelity check within the modelling
hierarchy, not independent validation. Twelve stations per loop were placed
along the seam and deliberately
clustered within $\pm4$\,mm of the corner, where the loops differ; all $24$
converged on the production grid with the field-quality criterion satisfied, no
melt reaching the outlet plane and no case at the boiling cap.

Table~\ref{tab:sil} reports the finite-volume-model assessment of these
commanded points.

\begin{table}[!htb]
  \centering
  \caption{Finite-volume assessment at $24$ commanded operating points.
    Station counts are not length fractions: the stations are clustered at the
    corner and therefore over-weight it by construction.}
  \label{tab:sil}
  \footnotesize
  \begin{tabular}{lcc}
    \toprule
     & open loop & closed loop \\
    \midrule
    Model root-side width (mm) & 2.010--3.678 & 1.722--2.088 \\
    Stations within band      & 9 of 12      & 12 of 12 \\
    Peak excursion (mm)       & 0.678        & 0.000 \\
    Surrogate MAE (mm)        & 0.089        & 0.049 \\
    Surrogate max error (mm)  & 0.247        & 0.188 \\
    Surrogate bias (mm)       & $+0.077$     & $+0.034$ \\
    \bottomrule
  \end{tabular}
\end{table}

Against the finite-volume predictions, all 12 closed-loop stations lie within
the illustrative root-side-width band, whereas 9 of 12 open-loop stations do so
and the maximum open-loop excursion is $0.678$\,mm. In the closed loop, the
surrogate MAE is $0.049$\,mm and its maximum error is $0.188$\,mm. The maximum
error is approximately $1/3.6$ of the open-loop excursion; the mean error is
approximately $1/13.8$ of that excursion. The positive bias means that the surrogate slightly
overpredicts root-side width and therefore tends to reduce heat input. This is
conservative only with respect to exceeding the upper root-side-width limit,
not with respect to other weld-quality criteria.

This is a quasi-steady comparison: the finite-volume model is run to steady
state at each fixed operating point and does not resolve the thermal lag of a
pool responding to a changing condition. The surrogate inherits the same
assumption because it was trained on steady-state model output. A fully
transient assessment would require the finite-volume model at every control
step, which at the measured mean of $467.6$\,s per finite-volume operating-point evaluation
and $217$ steps is approximately $28$\,h for a $4.4$\,s weld.

\begin{figure*}[!t]
  \centering
  \includegraphics[width=0.82\textwidth]{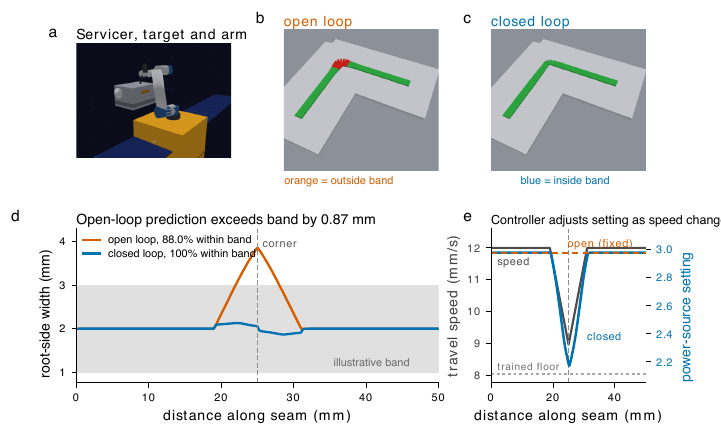}
  \caption{Model-based closed-loop study for an L-shaped seam in $1$\,mm 316L
    at the assumed microgravity/vacuum reference condition. (a) Illustrative
    servicer, target and UR5 cell; the robot uses the published kinematic
    parameters. (b, c) Simulated path coloured by the illustrative
    $1.0$--$3.0$\,mm root-side-width band. (d) Open- and closed-loop predictions.
    (e) Travel speed and commanded power-source setting. The two loops use the
    same trajectory; slowing from $12.0$ to $8.94$\,mm\,s$^{-1}$ raises heat input
    per unit length by $34\,\%$ at fixed setting, whereas control limits the rise
    to $5.5\,\%$ and keeps the surrogate prediction inside the band.}
  \label{fig:robot_demo}
\end{figure*}

Figure~\ref{fig:robot_demo} presents the seam-level open- and closed-loop
predictions, motion profile and power-source commands.

\section{Model-based sensitivity and transfer limits}
\label{sec:derisking}

\begin{figure*}[!t]
  \centering
  \includegraphics[width=0.90\textwidth]{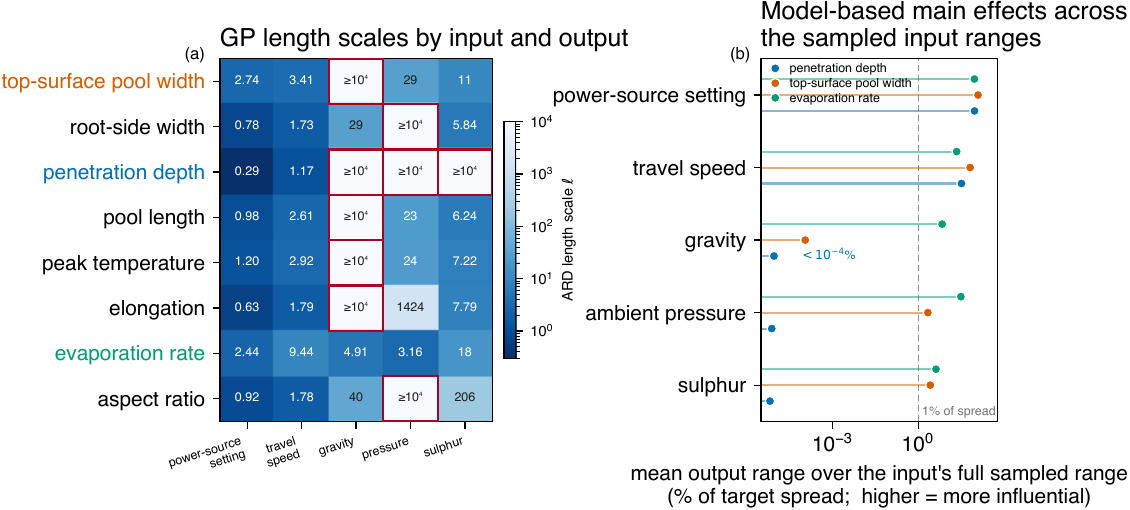}
  \caption{Model-based sensitivity of the GP surrogate. (a) Length scales fitted
    by automatic relevance determination after input standardisation;
    larger values indicate weaker response within the sampled domain. Values at
    the upper bound are marked. (b) Mean predicted output range obtained by
    varying one input over 40 levels across its sampled range and averaging over
    300 uniformly drawn background points (random seed 0), normalised by the
    corresponding output range. The pressure input represents effective local
    pressure, not chamber pressure. The diagnostic
    describes the reduced model and should not be interpreted as experimental
    process sensitivity.}
  \label{fig:model_sensitivity}
\end{figure*}

\begin{figure*}[!t]
  \centering
  \includegraphics[width=0.90\textwidth]{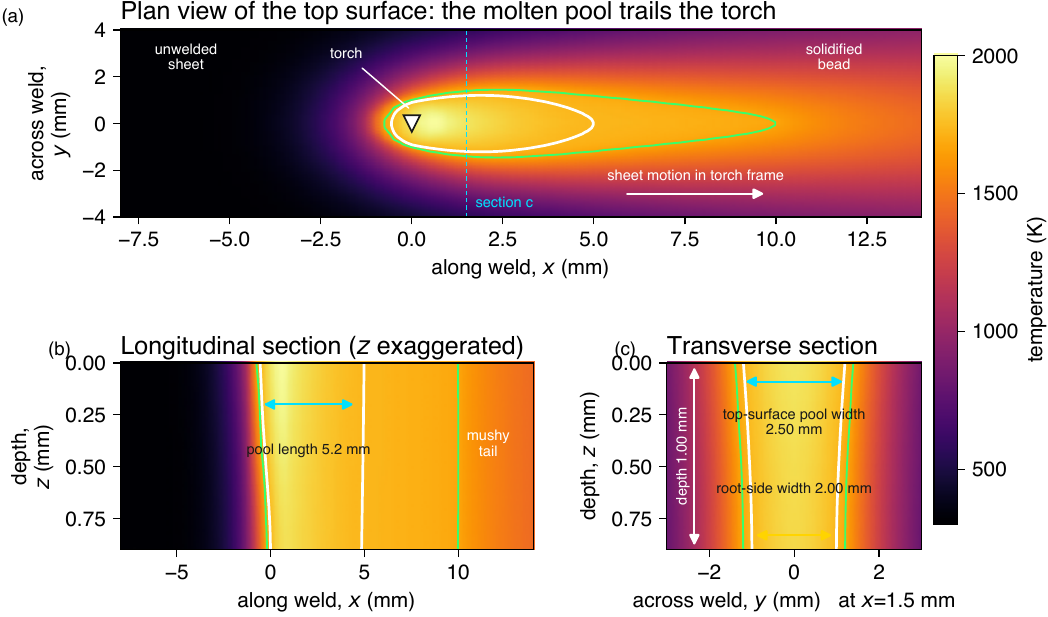}
  \caption{Finite-volume temperature field for the reference calculation.
    (a) Top-surface plan view; (b) longitudinal section; and (c) transverse
    section at the maximum surface width. The solidus and liquidus contours
    distinguish the mushy region from the liquid weld pool. Dimensions annotate
    the liquidus contour: pool length $5.2$\,mm, top-surface width $2.50$\,mm
    and root-side width $2.00$\,mm. The depth is limited by the 1\,mm sheet
    thickness. These are modelled liquidus-contour dimensions, not measurements
    of a solidified weld bead.}
  \label{fig:reference_pool}
\end{figure*}

Figures~\ref{fig:model_sensitivity} and \ref{fig:reference_pool} present the
surrogate sensitivity diagnostics and the corresponding finite-volume
temperature field. Both describe the implemented reduced model, not
experimental process behaviour.

Figure~\ref{fig:model_sensitivity} reports two complementary sensitivity
summaries. Panel (a) gives the GP length scales fitted by automatic relevance
determination after input standardisation; a larger value indicates a weaker
response within the sampled domain. Panel (b) varies one input across its
sampled range while averaging the predicted output range over 300 uniformly
drawn background points. This is a global main-effect diagnostic of the fitted
surrogate, not a causal sensitivity measure for the physical welding process.

The synergic power-source setting and travel speed have the shortest length
scales and the largest predicted influence on weld-pool dimensions. The gravity
length scale reaches its upper bound for most thermal and geometric outputs, and
the predicted change in penetration depth across the sampled gravity coordinate
is less than $10^{-4}$\,\% of its sampled output range. This is the expected numerical
invariance check because gravity is inactive in the thermal equations, not a
physical sensitivity result. The imposed local-pressure parameter acts most
strongly on evaporation rate through the assumed surface-loss relations, while
sulphur concentration influences the surface-tension-dependent geometric
responses.

These results must be interpreted within the governing model. Gravity enters
only the hydrostatic-to-capillary pressure ratio, and the model excludes momentum,
free-surface deformation, gravity-dependent droplet detachment and arc
behaviour. The invariance of the eight thermal outputs therefore does not
establish that a physical weld pool is gravity insensitive. Likewise, transfer
to GTAW, other GMAW transfer modes, laser welding or electron-beam welding would
require process-specific heat-source models, data and validation.


\section{Sensing interfaces}
\label{sec:sensing}

The framework is designed to accept measurement updates, but the present study
does not test this interface with hardware. The following channels are
therefore candidate requirements for a future hardware programme, not
experimental results or work undertaken in this study.

\textit{Welding current and arc voltage} can be acquired from the power source
at kilohertz sampling rates. Their product gives electrical power, while arc
efficiency requires a calorimetric or thermal estimate.

\textit{Thermal imaging} can estimate the top-surface temperature field and
thereby inform heat-source distribution parameters
\cite{fernandez2020embedded}. Quantitative comparison requires
radiometric calibration, an emissivity model and treatment of arc radiation.

\textit{Vision} of the weld pool and solidified bead can provide geometric
observations and is an established basis for closed-loop weld control
\cite{chen2009closed}. For thin sheet, a root-side
view would additionally measure root-side width and the onset of burn-through.

In the Kennedy--O'Hagan structure of Section~\ref{sec:priors}, each channel enters
as an observation with its own error model, updating the parameter posteriors
between welds. Within-weld estimation would additionally require an observation
model, an estimator and a transient process model at the control rate; forward
query latency alone does not establish that capability. Synchronised sensor
records, model versions and parameter estimates can provide traceability, but
inspection and acceptance remain outside the present model-based study.


\section{Future testing and model development}
\label{sec:roadmap}
The present study is limited to modelling and simulation. Future experiments
should assess predictive accuracy and the performance of the coupled robotic
system. The work proposed below would support model validation; weld-procedure
qualification would require separate acceptance criteria based on structural
requirements and applicable welding standards. The illustrative root-side-width
band in Section~\ref{sec:robot} is not a qualification criterion.

\textit{Bench trials at 1\,g and atmospheric pressure.} Measurements
of top and root bead geometry, thermal response and electrical signals should be
collected across the power-source settings. These data would update the parameter
distributions and model-discrepancy term in Section~\ref{sec:priors}; a separate
validation set would assess predictive error and interval coverage.

\textit{Thermal-vacuum testing at 1\,g.} Hardware testing should prioritise
the effects of pressure and thermal boundary conditions. Chamber pressure,
radiative-sink temperature and the initial
coupon and fixture temperatures should be varied in a controlled sequence so
that their effects can be separated. Measurements of heat rejection, local gas
pressure and composition, evaporation-related mass loss, arc behaviour,
penetration and bead geometry would test the pressure- and temperature-dependent
assumptions used by the model.

\textit{Targeted reduced-gravity testing.} Parabolic flight provides seconds of
microgravity per parabola \cite{mcauley2026laser}. The
measurements should prioritise the unresolved gravity-dependent mechanisms:
free-surface deformation, metal transfer, and weld-pool and final-bead geometry.
The thermal model is invariant to gravity because gravity does not enter its
energy equation. This does not establish that the physical process is gravity
insensitive. Reduced-gravity trials would assess the mechanisms omitted from the
model and the relevance of the separate hydrostatic-to-capillary pressure ratio.
Thermal-vacuum testing remains the immediate priority for assessing the
pressure-dependent assumptions used here.

\textit{Hardware-in-the-loop trials.} The deployed
surrogate and controller should be connected to a robotic
manipulator, welding end-effector, sensing chain and representative target
fixture. Hardware-in-the-loop (HIL) trials should assess seam tracking,
contact-tip-to-work distance, arm and target compliance, command and sensing
latency, disturbances, off-nominal events and safe shutdown. The tests should
progress from non-welding motion and sensor emulation to integrated welding in
the thermal-vacuum environment.

\Needspace{3\baselineskip}
\textit{Model development and traceability.} Free-surface deformation is a
priority for extending the model. Further extensions should address droplet
detachment, the CMT current and wire-motion cycle, robot--structure dynamics and
solidification microstructure. Each extension requires numerical verification
and experimental validation. Test records should link the model version,
parameter estimates, sensor measurements and weld observations so that changes
in predictive performance can be traced.

\section{Conclusions}

This paper combines a reduced thermal weld-pool model, data-driven surrogates and
a robotic-control simulation for screening candidate in-space arc-welding
conditions. In repeated cross-validation, the GP is more accurate than the neural
network for all eight outputs. Across ten deterministic boundary cases, 92.5\,\%
of finite-volume responses fall within nominal 90\,\% GP intervals. A three-output
query takes 249\,$\mu$s, corresponding to a measured speed-up of
$1.6\times10^{6}$ relative to the median finite-volume calculation.

In the L-seam study, all 12 closed-loop operating points assessed with the
finite-volume model satisfy the illustrative root-side-width band, compared with 9
of 12 open-loop points. This supports the feasibility of the simulated control
strategy, but does not validate predictions for physical welds. Power-source setting and travel
speed dominate the model's geometric responses, whereas the imposed local-
pressure parameter affects predicted evaporation through the assumed loss
relations. The thermal outputs are invariant to the sampled gravity coordinate
because gravity is absent from the implemented energy equation; no physical
gravity-insensitivity claim follows while momentum, free-surface deformation,
arc behaviour and droplet transfer remain unresolved. Further development
requires
formal grid convergence, experimental calibration and independent coupon
validation, thermal-vacuum and targeted reduced-gravity testing, HIL evaluation
with a robotic manipulator, and transient free-surface and
metal-transfer extensions.

\section*{Acknowledgements}
\addcontentsline{toc}{section}{Acknowledgements}
This work was supported by the UK Space Agency under National Space Innovation
Programme 2 (NSIP 2) through the Intelligent SPace Arc-welding Robotic Kit
(ISPARK) project. The corresponding author thanks colleagues at Space Park
Leicester and TWI Ltd. for constructive discussions and the project engineers
for their support. Generative-AI tools assisted with code debugging, language
editing, page-limit management, consistency checking and review. The authors
reviewed and verified the technical content and conclusions and accept
responsibility for the manuscript.

{\fontsize{8.5}{8.8}\selectfont
\bibliographystyle{ieeetr}
\bibliography{references}
}

\end{document}